%% file: main.tex
\documentclass{article}

\usepackage{arxiv}

\usepackage[utf8]{inputenc} % allow utf-8 input
\usepackage[T1]{fontenc}    % use 8-bit T1 fonts
\usepackage{hyperref}  % hyperlinks
\usepackage{url}            % simple URL typesetting
\usepackage{booktabs}       % professional-quality tables
\usepackage{amsfonts}       % blackboard math symbols
\usepackage{amsmath}
\usepackage{cleveref}
\usepackage{amssymb}
\usepackage{mathtools}
\usepackage{nicefrac}       % compact symbols for 1/2, etc.
\usepackage{microtype}      % microtypography
\usepackage{graphicx}
\usepackage{diagbox}
\usepackage{doi}
\usepackage{color}
\usepackage{thmtools}
\usepackage{caption}
\usepackage{subcaption}
\usepackage[ruled,linesnumbered,noend]{algorithm2e}
\usepackage{floatrow}
\newfloatcommand{capbtabbox}{table}[][\FBwidth]

\newtheorem{problem}{Problem}
\newtheorem{definition}{Definition}

\newcommand{\collabdm}{\textsf{CollabDM}\xspace}

\title{One-shot Collaborative Data Distillation}
\author{\normalfont William Holland\thanks{Corresponding Author. Email: rayne.holland@data61.csiro.au.}, Chandra Thapa, Sarah Ali Siddiqui, Wei Shao, and Seyit Camtepe \\ CSIRO Data61, Sydney, Australia}

\renewcommand{\shorttitle}{\textit{One-Shot Collaborative Data Distillation}}

\hypersetup{
pdftitle={One-Shot Collaborative Data Distillation},
pdfsubject={q-bio.NC, q-bio.QM},
pdfauthor={William L.~Holland},
pdfkeywords={Data Distillation, Federated Learning},
}

\begin{document}
\maketitle

\begin{abstract}

Large machine-learning training datasets can be \textit{distilled} into small collections of informative synthetic data samples.
These synthetic sets support efficient model learning and reduce the communication cost of data sharing.
Thus, high-fidelity distilled data can support the efficient deployment of machine learning applications in distributed network environments.
A naive way to construct a synthetic set in a distributed environment is to allow each client to perform local data distillation and to merge local distillations at a central server.
However, the quality of the resulting set is impaired by heterogeneity in the distributions of the local data held by clients.
To overcome this challenge, we introduce the first \emph{collaborative} data distillation technique, called \textsf{CollabDM}, which captures the global distribution of the data and requires only a single round of communication between client and server.
Our method outperforms the state-of-the-art one-shot learning method on skewed data in distributed learning environments.
We also show the promising practical benefits of our method when applied to attack detection in 5G networks.

\end{abstract}

\section{Introduction}

\input{sections/01_introduction}

\section{Related Work}

\input{sections/02_related_work}

\section{Preliminaries}

\input{sections/03_prelims}

\section{Collaborative Data Distillation}

\input{sections/04_solution}

\section{Experiments}

\input{sections/05_experiments}

\section{Conclusion}

\input{sections/06_conclusion}

\section{Acknowledgments}
This research work is partially conducted as part of the 6G Security Research and Development Project, as led by the Commonwealth Scientific and Industrial Research Organisation (CSIRO) through funding appropriated by the Australian Government’s Department of Home Affairs. This paper does not reflect any Australian Government policy position. For more information regarding this Project, please refer to \url{https://research.csiro.au/6gsecurity/}.

\appendix

\bibliographystyle{acm}
\bibliography{bib}

\end{document}

%% file: sections/01_introduction.tex
Machine learning models trained on massive datasets are susceptible to high training times, slow research iteration, and poor eco-sustainability.
To overcome these problems, and to increase the scalability of machine learning applications, large datasets can be \textit{distilled} into small collections of informative synthetic data samples \cite{wang2018dataset}.
If the distilled data effectively captures the original dataset, machine learning models can be trained \textit{efficiently} on the synthetic data with accuracy comparable to models trained on the original data.

In addition to computational efficiency, data distillation has the benefit of both reducing the communication cost of data sharing and, as only synthetic samples are shared, providing privacy to data owners \cite{chen2022private}.
These benefits are notable in applications such as 5G mobile networks, where massive volumes of data are generated from diverse sets of sources.
In this setting, distilled data can be shared, in a safe and communication efficient manner, across heterogeneous domains and utilized for robust model training.

However, distributed learning is impaired by heterogeneity between the local distributions of the data held by clients \cite{li2020federated,li2021fedbn}. 
Further, sharing locally distilled datasets for global model training can amplify the perverse impacts of data heterogeneity \cite{huang2023federated}. 
This motivates the creation of data distillation techniques that synthesize a \textit{global} synthetic dataset through the collaboration of clients. 
Collaboration allows diverse data sources to participate in a global distillation process without sharing local data.
The resulting global synthetic set can be shared across parties and utilized for applications such as neural network architecture search \cite{zhao2021dataset}, (global) model training, and continual learning \cite{rosasco2021distilled}. 

Standard data distillation techniques \cite{cazenavette2022dataset,wang2018dataset, zhao2020dataset} operate in a centralized and static model in which the whole dataset is available in a single location.
Adapting these methods (efficiently) in a distributed learning environment is non-trivial and remains an open challenge.
For example, Pi \textit{et al.} propose a federated learning framework that performs global data distillation \cite{pi2023dynafed}.
The synthetic data is optimized such that the parameter trajectory of the model trained on it matches the parameter trajectory of the model trained through standard federated learning.
However, this procedure operates on a \textit{single} model initialization and distillation algorithms that match training trajectories typically generate training trajectories across a large number of random model initializations \cite{cazenavette2022dataset}.
Thus, the authors only implement a fraction of the full algorithm.

The challenge with adapting data distillation techniques to collaborative settings is that most involve many iterations of model training, which would, thus, involve many rounds of communication and parameter sharing. 
Consequently, they incur large communication overheads in a distributed setting and negate many of the benefits they promise.
To overcome this limitation, we present a collaborative data distillation algorithm based on \textit{distribution matching} \cite{wang2022cafe,zhao2023dataset}.
In distribution matching \textit{no model training is required}.
Instead, the synthetic data are optimized to match the distribution of real data in a family of {random} lower-dimensional embedding spaces.
As the embedding spaces are randomly initialized, they can be distributed to clients with random seeds, mitigating the communication burden of transmitting and training model parameters.

Further, in distribution matching, the mean of the embeddings on real data is required to compute the loss functions for synthetic data optimization.
Thus, with the random seeds for these embeddings initialized a priori, the clients can compute \textit{all} means (one for each iteration of synthetic data training) in a single batch and transmit them to the server in a \textit{single-round} of communication.
Consequently, synthetic data is distilled collaboratively with a small communication overhead.

Prior work has utilized data distillation for communication reduction in federated learning scenarios \cite{goetz2020federated, hu2022fedsynth, liu2022meta, song2023federated, xiong2023feddm,  zhou2020distilled}.
Here, clients distill their data \textit{independently} of each other and upload the distilled data to the server.
The global model is then updated with the information distilled in the synthetic data.
For large models, the distilled data is often more compact than the model parameters.
Therefore, synthetic data can offer lower per-round communication and faster model convergence than standard approaches that aggregate local model parameters into a global model.
These methods provide a heuristic for reducing the communication cost of federated learning.
However, they do not optimize the synthetic data over the \textit{global} data distribution.
The significance of a global synthetic dataset is that it provides distributed settings with efficient methods for additional applications, such as neural architecture search  and continual learning. 

\paragraph*{Motivating Application}
To motivate the field of collaborative data distillation, we provide a target application for the technique: 5G mobile networks.
Next-generation mobile networks are built on edge networks, where network resources are placed close to end-users and are often spread across multiple tenants and domains.
This creates a landscape where large volumes of data are generated at a growing number of locations, often within specified trust boundaries.
The generated data can be used for a burgeoning range of 5G machine learning applications \cite{kaur2021machine}.
The data can be heterogeneous, motivating the need for globally trained models that generalize well.
However, the generated data can be both large in size and private, preventing its transmission to a central point that orchestrates the machine learning applications.
This challenge can be overcome with a compact global synthetic set, which can be easily shared among edge networks to support the relevant machine learning applications.

In our studies, along with standard benchmark datasets, we have considered attack detection in network traffic.
In this setting, traffic at different points in the network can be monitored by a device with a general CPU or GPU.  
The device maintains a neural network to classify incoming traffic as benign or anomalous.
If multiple points in the network contribute to training a global synthetic set, robust model training can be performed to capture the global dynamics of data generated across the network.

\paragraph*{Contributions}

\begin{itemize}
    \item We provide the first distributed data distillation algorithm, \collabdm, that captures the dynamics of the global data distribution in a \textit{single-round} of communication.
    \item The algorithm is tested against benchmark datasets.
    Results indicate that our technique outperforms the state-of-the-art one-shot learning method DENSE \cite{zhang2022dense} on heterogeneous data partitions.
    The global synthetic sets generated by \collabdm are remarkably robust to the underlying data distribution, with only very small reductions in performance when the level of skew in the data distribution increases.
    \item The algorithm is tested in a target distributed learning environment: 5G networks. 
    This represents a new application for data distillation techniques.  
    Results demonstrate that data distillation provides a promising direction for supporting machine learning applications in 5G networks.
\end{itemize}

%% file: sections/02_related_work.tex
\paragraph*{Data Distillation}

Data distillation methods aim to synthesize small and high-fidelity data summaries, which distill the most important information from a target dataset \cite{sachdeva2023data}.
The summaries can serve as effective drop-in replacements for the original dataset in machine-learning applications.
Data distillation methods can be categorized into three types: meta-learning, parameter matching, and distribution matching. 

Meta-learning techniques \cite{nguyen2020dataset,wang2018dataset} aim to minimize the expected loss incurred on real data for models trained on the synthetic set.
This involves a bi-level optimization, where an inner loop trains a model with respect to the synthetic dataset, and the outer loop updates the synthetic set (considered as a hyperparameter) based on the loss observed on the model by real data.
Parameter matching techniques allow the synthetic data to imitate the influence of the target dataset on model training.
For example, synthetic data can be distilled to match the training gradients \cite{zhao2020dataset} or parameter trajectories \cite{cazenavette2022dataset} observed during training on real data.

In distribution matching \cite{wang2022cafe,zhao2023dataset}, the synthetic data are optimized to match the distribution of real data in a family of lower-dimensional embedding spaces.
In contrast to prior approaches, distribution matching involves a single-level optimization.
It is, therefore, considered less computationally intensive and more scalable.

\paragraph*{Virtual Learning}

Federated learning involves building a local surrogate function to approximate the local training objective.
By sending local surrogate functions to the server, the server can build a global surrogate around the current solution.
The aim is to build local surrogates that are informative and succinct.
Local synthetic data can be constructed to capture information about the local update at the client and build local surrogate functions \cite{xiong2023feddm}.
For example, locally distilled data can be used to approximate gradient updates \cite{goetz2020federated, liu2022meta}, 
minimize the difference between models trained on real and synthetic data \cite{hu2022fedsynth} or
communicate local approximations in the loss landscape \cite{wang2020federated}.

Huang \textit{et al.} propose an iterative method that utilizes local and global distillation \cite{huang2023federated}.
They iteratively refine local and global synthetic data.
The global virtual data is used as an anchor on the server side for model training.
Similarly, Liu \textit{et al.} attempt to distill synthetic data with global dynamics \cite{liu2023few}.
The distilled data is optimized to mimic the parameter trajectories of the global model under the standard FedAvg~\cite{mcmahan2017communication} algorithm. 
The authors observe that the updated dynamics of the global model contain knowledge about the global data distribution.
This knowledge is transferred to a synthetic dataset at the server.
However, data distillation with trajectory matching typically requires training on lots of randomly initialized models. Therefore, the actual data distillation algorithm is only partially implemented.

Note that the above methods are all \textit{multi}-shot; that is, they require multiple rounds of communication.

\paragraph*{One Shot Federated Learning}

One-shot federated learning involves completing a federated learning objective in a single round of communication. 
Single-round communication is in high demand for practical applications \cite{su2023one} and has advantages such as reducing the risk of being attacked \cite{zhang2022dense}. 
Most one-shot federated learning methods are based on knowledge distillation \cite{hinton2015distilling} or data distillation \cite{wang2018dataset}.

Methods based on knowledge distillation utilize the local models as teachers for the global model \cite{guha2019one,zhang2022dense}.
Guha \textit{et al.} propose a method where each client trains a model to completion and ensemble methods are used to train a global model \cite{guha2019one}.
This approach involves a public dataset.
Zhang \textit{et al.} propose a two-stage method that trains a global model through a data generation stage and a model distillation stage \cite{zhang2022dense}.
The first stage uses ensemble models obtained from clients to train a global data generator.
The knowledge from ensemble models is distilled in the data generator and used to train a global model.

For methods based on data distillation, clients distill synthetic data locally (and independently of each other) and send the summaries to the server \cite{song2023federated, zhou2020distilled}, constituting a single round of communication.
The server then trains the model on aggregated synthetic data.
Our method adopts this template.
However, our approach differs in that clients send additional computations, allowing the server to refine the synthetic data according to a \textit{global} loss function.
Thus, our approach is able to better combat data heterogeneity observed across the clients.

%% file: sections/03_prelims.tex
The first part of this section covers key notation and the problem definition.
The second part introduces the main data distillation frameworks \cite{lei2023comprehensive}: meta-learning, parameter matching, and distribution matching.
The meta-learning and parameter-matching frameworks will help demonstrate the challenges of distributed data distillation.
Our approach is based on the distribution matching framework, which supports a collaborative algorithm that overcomes these challenges.

\subsection{Notation}
Let $\mathcal{D} \overset{\Delta}{=} \{(x_i, y_i)\}_{i=1}^{|\mathcal{D}|}$ be the data set that needs to be distilled, where $x_i\in \mathcal{X}$ denotes the input features and $y_i\in\mathcal{Y}$ is the label for $x_i$.
Throughout, the notation $d \sim \mathcal{D}$ refers to a data point $d$ selected uniformly at random from the set $\mathcal{D}$.
Given a data budget $n \in \mathrm{Z}^{+}$, a data distillation technique aims to synthesize a high-fidelity summary 
$\mathcal{S} \overset{\Delta}{=} \{(\tilde{x_i}, \tilde{y_i})\}_{i=1}^{n}$ such that $n\ll |\mathcal{D}|$.
The small distilled dataset should achieve a comparable generalization performance to the large original dataset.

For a given learning algorithm $\Phi_\theta:\mathcal{X}\rightarrow \mathcal{Y}$, with parameterization $\theta$, the empirical risk $\mathcal{R}$ on parameterization $\theta$ and input data $\mathcal{D}$ is defined as 
\begin{align*}
    \mathcal{R}(\mathcal{D};\theta) &= \sum_{i=1}^{|\mathcal{D}|} l(\Phi_{\theta}(x_i),y_i), 
\end{align*}
where $l$ is a loss function.
A training algorithm for $\Phi$ attempts to find $\theta$ that minimizes $\mathcal{R}$.

\subsection{Problem Definition}

\begin{definition}[Data Distillation (\cite{sachdeva2023data})]
    Given a learning algorithm $\Phi$, let $\theta^{\mathcal{D}}$, $\theta^{\mathcal{S}}$ represent the optimal set of parameters for $\Phi$ on, respectively, $\mathcal{D}$ and $\mathcal{S}$. \emph{Data distillation} is defined as the optimization of the following:
    \begin{align}
        \arg \min_{\mathcal{S}} \left(  \sup \{\hspace{2mm} |l(\Phi_{\theta^{\mathcal{D}}}(x),y)-l(\Phi_{\theta^{\mathcal{S}}}(x),y)| \hspace{2mm} \}_{\substack{x \sim \mathcal{X} \\ y\sim \mathcal{Y}}} \right).
        \label{eqn:data_distill}
    \end{align}
\end{definition}
Thus, the objective is to \textit{extract} the knowledge from $\mathcal{D}$ and transfer it to the synthetic set $\mathcal{S}$, such that the model trained on $\mathcal{S}$ should achieve comparable generalization to the model trained on $\mathcal{D}$.

\begin{problem}[Collaborative Data Distillation]
    The dataset $\mathcal{D}$ is split over $K$ disjoint clients that can communicate with a central server.
    Let $\mathcal{D}_i$ be the data stored at client $i$.
    For $\mathcal{D} = \cup_{i=1}^K \mathcal{D}_i$, \emph{collaborative data distillation} aims to solve the objective of Equation \eqref{eqn:data_distill} under the conditions that \begin{enumerate}
        \item The server cannot observe $\mathcal{D}$.
        \item Client $i$ cannot observe $\mathcal{D} \setminus \mathcal{D}_i$.
    \end{enumerate} 
    \label{prob:cdd}
\end{problem}

With compact synthetic sets, collaborative data distillation aims to reduce the communication overhead in distributed machine-learning applications at a minimal cost in terms of fidelity.

\subsection{Data Distillation with Meta-Learning}

Meta-learning based methods \cite{bohdal2020flexible,sucholutsky2021soft,wang2018dataset, zhou2022dataset} treat $\mathcal{S}$ as a hyperparameter, which is updated by a meta (outer) algorithm and a base (inner) algorithm solves a conventional learning problem with respect to the synthetic dataset. 
The objective can thus be formulated as a bi-level optimization:
\begin{align}
    \mathcal{S}^* = \arg \min_{\mathcal{S}} \mathcal{R}(\mathcal{D};\theta^{\mathcal{S}})
    \label{eqn:meta_learning_DD}
\end{align}
subject to 
\begin{align*}
    \theta^{\mathcal{S}} = \arg \min_{\theta} \mathcal{R}(\mathcal{S};\theta).
\end{align*}
The inner loop, which optimizes parameters on the synthetic data, can be realized through gradient descent or kernel regression.
The objective function can be defined as the meta-loss $\mathcal{L}(\mathcal{S}) = \mathcal{R}(\mathcal{D},\theta^{\mathcal{S}})$.
Consequently, the synthetic data can be updated as $\mathcal{S} = \mathcal{S}- \alpha \nabla_{\mathcal{S}} \mathcal{L}(\mathcal{S})$ for learning rate $\alpha$. 

\subsection{Data Distillation with Parameter Matching}
Data matching \cite{cazenavette2022dataset, zhao2020dataset} aims to align the byproducts of model training on real and synthetic data.
The synthetic data learns to \textit{mimic} the influence of real data on model training. 
The objective function of data matching can be summarized as follows:
\begin{align*}
    \mathcal{L}(\mathcal{S}) = \sum_{k=0}^T Q(\phi(\mathcal{D}, \theta^{(k)}), \phi(\mathcal{S}, \theta^{(k)}))
\end{align*}
subject to 
\begin{align*}
    \theta^{(k+1)} = \theta^{(k)} - \eta\nabla_{\theta^{(k)}} \mathcal{R}(\mathcal{S};\theta^{(k)}),
\end{align*}
where $Q$ is a distance function, $\eta$ is the learning rate, and $\phi$ maps a dataset to informative spaces such as gradient, parameter, and feature spaces.
For example, the map $\phi(\mathcal{S}, \theta) = \nabla_{\theta^{}} \mathcal{R}(\mathcal{S};\theta^{}) $ \cite{zhao2020dataset} equates to matching the gradients, with respect to $\theta$, of the observed empirical risk, and $\mathcal{S}$ is optimized to mimic the gradients observed on $\mathcal{D}$ during model training.
In practice, the full dataset might be replaced with a batch to save memory and facilitate faster convergence.

\subsection{Data Distillation with Distribution Matching}

The underlying bi-level optimization of prior approaches is often expensive in terms of computation and memory. 
To mitigate these costs, distribution matching \cite{wang2022cafe,zhao2023dataset} aims to solve a correlated proxy task that restricts optimization to a single level and improves scalability.
Instead of matching the quality of the models generated by $\mathcal{D}$ and $\mathcal{S}$, distribution matching attempts to match the underlying distributions of $\mathcal{D}$ and $\mathcal{S}$. 
The assumption here is that datasets with the same distribution will lead to similarly trained models.

Distribution matching uses a collection of random parametric encoders to embed data into low-dimensional latent spaces.
Distance metrics can then be used to compute the distribution mismatch between $\mathcal{D}$ and $\mathcal{S}$.
Formally, given a set of encoders $\mathcal{E} = \{ \psi_i: \mathcal{X} \rightarrow \mathcal{X}_i\}$, the optimization objective, under maximum mean discrepancy, is:
\begin{align*}
    \arg \min_{\mathcal{S}}  \mathrm{E}_{\substack{\psi \sim \mathcal{E} \\ y\sim \mathcal{Y}}}\left[ \lVert \mathrm{E}_{\substack{x \sim \mathcal{D}^y}}[ \psi(x) ] - \mathrm{E}_{\substack{x\sim \mathcal{S}^y}}[\psi(x) ] \rVert^2  \right],
\end{align*}
where $\mathcal{D}^y = \{ x \mid (x,y) \in \mathcal{D}\}$.
This objective, for a given $\psi \in \mathcal{E}$, can be approximated with the following empirical loss:
\begin{align}
    \mathcal{L} =\sum_{y \in \mathcal{Y}} \left\lVert \frac{1}{|D^y|} \sum_{x\in D^y} \psi_{\theta} (x) - \frac{1}{|S^y |}  \sum_{x \in S^y } \psi_{\theta} (x)  \right\rVert^2,
    \label{eqn:DM_loss}
\end{align}
where $D^y\subset \mathcal{D}^y$ denotes a batch of real data and $S^y \subset \mathcal{S}^y $ denotes a batch of synthetic data.
Typically $\mathcal{E}$ is generated randomly and each $\psi \in \mathcal{E}$ has the same network architecture.

%% file: sections/04_solution.tex
In the collaborative setting, the dataset $\mathcal{D} = \cup_{i=1}^{K}\mathcal{D}_i $ is split over $K$ disjoint clients that can communicate with a central server.
The goal of collaborative data distillation is to produce a synthetic dataset $\mathcal{S}$ at the server that achieves similar generalization performance to $\cup_{i=1}^{K}\mathcal{D}_i $.
As a starting point, a straightforward solution would be to get each client $i$ to independently distill their own synthetic dataset $\mathcal{S}_i$ (using any data distillation method) and set $\mathcal{S} = \cup_{i=1}^K \mathcal{S}_i$.
However, under the influence of data heterogeneity, the locally distilled data could be biased and, consequently, produce a distillation that does not capture the \textit{global} data distribution.

Alternatively, the global dynamics of the data could be captured by adapting a full data distillation algorithm to a federated learning setting.
In the following subsection, we provide a framework for the adaptation of the meta-learning and data-matching algorithms.
This framework will act as a strawman solution to our collaborative distillation algorithm based on distribution matching.

\subsection{Strawman Collaborative Distillation}

The collaborative distillation process begins with the server initializing a set of synthetic data $\mathcal{S}$.
This can be achieved with a random initialization or by asking clients to transmit local data distillations: $\mathcal{S}= \cup_{i=1}^K \mathcal{S}_i$.
With the synthetic data initialized, it is then updated iteratively.
For the \textit{meta-learning} and \textit{data-matching} frameworks, each iteration has the following steps:
\begin{enumerate}
    \item The server \textit{retrieves} a network $\theta$.
    The network parameters could be generated randomly or retrieved from a cache.
    \item The network $\theta $ is \textit{updated}.
    The update could be based on either $\mathcal{R}(\mathcal{D}; \theta)$ or $\mathcal{R}(\mathcal{S}; \theta)$.
    For the former, clients conduct federated learning.
    The latter objective can be executed on the server.
    \item $\theta$ is broadcast to clients.
    \item The loss function $\mathcal{L}(\mathcal{S})$ is computed.
    The clients collaborate to compute $\Phi_{\theta}(D)$, for batch $D \subset \mathcal{D}$.
    \item Synthetic data is updated based on the gradient of the loss function.
\end{enumerate}

At each iteration, model parameters are broadcast to clients, and clients send losses incurred on real data to the server.
For large models, this can create a communication overhead that compromises the benefits of collaborative distillation.
In other words, similar to federated learning, the framework involves multiple rounds of communication that broadcast model parameters to clients.

\begin{figure}
    \centering
    \includegraphics[width=0.6\linewidth]{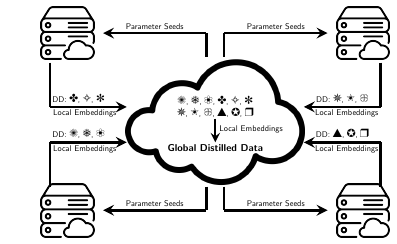}
    \caption{Overview of \collabdm.
    In a single round of communication, the server sends seeds to initialize learning models.
    The client then distills local data and computes embeddings on the seeded models. 
    Locally distilled data and computed embeddings are then sent to the server.
    The server uses the embeddings to refine the distilled data to reflect the global data distribution.}
    \label{fig:collabdm}
\end{figure}

This limitation can be overcome with distribution matching.
As distribution matching does not require model training, Step 2 from the framework can be removed.
Further, as embeddings are initialized randomly, they can be broadcast with a random seed.
Thus, distribution matching can be implemented in a collaborative learning environment without explicitly sharing network parameters. 

\subsection{Collaborative Distribution Matching}

\begin{algorithm}[t]
    \SetAlgoLined
    \DontPrintSemicolon
    \SetKwProg{myproc}{define}{}{}
    \KwIn{Number of clients $K$, number of global iterations $T$, proportion  $\varepsilon$ of clients participating in each iteration, learning rate $\eta$ and batch size $B$ for real data}
    \KwOut{Global distilled dataset $\mathcal{S}$}
    \myproc{\textup{$\textsf{ServerDM}()$}}
    {
        \For{$t \in \{1,\ldots, T\}$}
        {
        $\alpha_t \gets $ random seed\;
        Generate random subset of clients $Z_t \subset [K]$ with $|Z_t| = \varepsilon K$\;
        }
        \For{$k \in \{1,\ldots, K\}$}
        {
            $A_k \gets \{\alpha_t| k \in Z_t\}$\; 
            $\mathcal{S}_k, L_k  \gets \textsf{ClientDM}(A_k)$
        }
        \tcp{Begin server data distillation}
        $\mathcal{S}\gets \cup_{k=1}^K \mathcal{S}_k$\;
        \For{$t \in \{1,\ldots, T\}$}
        {
        \For{$y\in \mathcal{Y}$}
        {
            Generate random batch $S_t^y \subset \mathcal{S}^y$ of synthetic data \;
            $ \mathcal{L}^y \gets \left\lVert \frac{1}{|\varepsilon K|} \sum\limits_{k\in Z_t} L_{t,k}^y - \frac{1}{|S_t^y |}  \sum\limits_{x \in S_t^y } \psi_{\alpha_t} (x)  \right\rVert^2$\;
        }
        \tcp{Compute loss according to Eq.~\eqref{eqn:data_distill}}
        $\mathcal{L} \gets \sum_{y \in \mathcal{Y}} \mathcal{L}^y $\;
        $\mathcal{S} \gets \mathcal{S} - \eta \nabla_{\mathcal{S}}\mathcal{L}$\;
        }
        \KwRet $\mathcal{S}$\;
    }
    \myproc{\textup{$\textsf{ClientDM}(A_k)$}}
    {
        $\mathcal{S}_k \gets $ Compute data distillation on $\mathcal{D}_k$\; 
        $L_k \gets \varnothing$\;
        \For{$\alpha_t \in A_k$}
        {
            \For{$y \in \mathcal{Y}$}
            {
                Generate random batch $D_{t,k}^y \subset \mathcal{D}_k^y$ of real data with $|D_{t,k}^y| = B$\;
                $L_{t,k}^y \gets \frac{1}{B} \sum_{x \in D_{t,k}^y} \psi_{\alpha_t} (x)$\;
            }
            $L_k \gets L_k \cup \{L_{t,k}^{y}\}_{y \in \mathcal{Y}}$\;
        }
        \KwRet $\mathcal{S}_k, L_k$\;
    }
\caption{Collaborative Distribution Matching (\collabdm)}
\label{alg:CollabDM}
\end{algorithm}

The goal of Collaborative Distribution Matching (\collabdm) is to compute the loss function in Equation~\eqref{eqn:DM_loss} for each embedding $\psi \in \mathcal{E}$.
The gradient of the loss is used to update the synthetic dataset stored on the server.
As the loss function is calculated over the global dataset $\mathcal{D}=\cup_{i=1}^K \mathcal{D}_i$, the updates are able to capture the global dynamics of the data. 
Equation~~\eqref{eqn:DM_loss} can be split into two components: the embeddings on real data, which are computed at clients, and the embeddings on synthetic data, which are computed at the server.
As $\mathcal{E}$ is fixed, it can be broadcast to clients prior to the distillation process.
This allows each client to compute the mean embeddings on real data, one for each iteration of server training, in a single batch and complete their share of the collaboration in a single round of communication.
A high-level overview of the procedure is presented in Figure~\ref{fig:collabdm}.
We now outline the full algorithm (provided in Algorithm \ref{alg:CollabDM}) in more detail.

Let $|\mathcal{E}| = T$ denote the number of rounds required to distill synthetic data through distribution matching.
In \collabdm, for each future training round $t\in \{1,\ldots,T\}$, the server begins by selecting a random seed $\alpha_t$ to encode a lower-dimension embedding and selecting a subset of clients $Z_t \subset \{1,\ldots, K\}$ to participate in the round.
A batch of seeds $A_k = \{\alpha_t \mid k \in Z_t\}$ is then broadcast to each client $k$.
Once the clients receive embedding seeds from the server, client training begins. 
Each client has two roles.
First, the client performs a \textit{local} data distillation to produce $\mathcal{S}_k$.
Any data distillation technique could be used.
Local distillations will be used to initialize the synthetic data at the server.
Second, the client computes their contribution to each objective function.
For each embedding $\alpha_t \in A_k$ and label $y\in\mathcal{Y}$, the client selects a batch of real data $D_{t,k}^y\subset \mathcal{D}_k^y$, of size $B = |D_{t,k}^y|$, and computes the mean of the embeddings on the batch:
\[
L_{t,k}^y = B^{-1}\sum_{x\in D_{t,k}^y} \psi_{\alpha_t} (x).
\]
The collection of sums
\[
L_k = \bigcup_{t:\alpha_t \in Z_k} \bigcup_{y\in \mathcal{Y}} L_{t,k}^y
\]
is then sent to the server, along with $\mathcal{S}_k$.
This concludes the client's role in \collabdm.
Thus, in a single round of communication, the client receives $A_k$ and, subsequently, transmits $(L_k, \mathcal{S}_k)$.

The server can now complete data distillation through distribution matching.
The synthetic data is initialized through the local distillations $\mathcal{S} = \cup_{k=1}^K \mathcal{S}_k$.
The server then iterates through the embeddings in $\mathcal{E}$.
For each embedding $\alpha_t$, using the client computations on real data $\cup_{k\in Z_t} L_{t,k}$, the loss function $\mathcal{L}$ of Equation~\eqref{eqn:data_distill} is evaluated.
The synthetic data is then updated with the gradient of the loss with respect to $\mathcal{S}$:
\[
\mathcal{S} = \mathcal{S} - \eta \nabla_{\mathcal{S}} \mathcal{L}.
\]
As the embeddings on real data are constant with respect to $\mathcal{S}$, the gradients can be computed at the server.
Therefore, a global data distillation is achieved without further communication with clients.

\subsection{Parameter Optimization}
\label{sec:opt}
There are a number of optimizations that can be applied to the distribution matching objective to improve the utility of the synthetic data  \cite{liu2022dataset,shin2023frequency,wang2022cafe,zhao2023improved}.
These optimizations can be adapted to \collabdm.
Notably, the synthetic data variables can be parameterized in a more efficient manner to allow for the distillation of more instances (for the same memory budget) and an enhanced representation of the real data.
To achieve this, we adopt a technique called \textit{partition and expand} \cite{zhao2023dataset}.
For partition parameter $l$, each image $s \in \mathcal{S}$ is partitioned into $l \times l$ mini-samples, and each mini-sample is then expanded to the input data dimensions using differentiable augmentation:
\begin{align*}
    s \xrightarrow{\text{partition}} 
    \begin{bmatrix} 
        s_{1,1} & \dots  & s_{1,l}\\
        \vdots & \ddots & \vdots\\
        s_{l,1} & \dots  & s_{l,l} 
    \end{bmatrix}
    \xrightarrow{\text{up-sample}}
    s^{\prime}_1, s^{\prime}_2, \ldots, s^{\prime}_l
\end{align*}
Thus, the number of features extracted from $\mathcal{S}$ is increased without changing the storage budget.

%% file: sections/05_experiments.tex
We now evaluate the classification performance of deep networks that are trained on the synthetic data generated by our method.
A key parameter in \collabdm is the number of global iterations $T$.
As $T$ increases, we would expect a higher fidelity synthetic set, as we are able to expose the loss function to a greater number of random models.
However, increasing $T$ also increases the bandwidth overhead of the algorithm, as clients are required to send more random embeddings.
Therefore, we are interested in the trade-off between classification accuracy and the amount of data transferred.

Experiments are split across two settings.
First, we evaluate our approach against standard benchmark image classification datasets.
This allows for robust comparison with existing art.
Second, we provide an evaluation for a target application: attack detection in 5G mobile networks.
This target application extends the use of data distillation techniques to network traffic data and provides further motivation for collaborative data distillation (Problem~\ref{prob:cdd}).
Programs\footnote{Code is available here: \url{https://github.com/rayneholland/CollabDM/tree/main}} were executed on a Dell laptop with Intel Core i5-8350U CPU, 8GB RAM, x64-based processor, and NVIDIA Quadro P5000M, 16 GB.

\subsection{Training and Evaluation Setup}

\begin{table}[]
    \centering
    \resizebox{0.7\columnwidth}{!}{%
    \begin{tabular}{|c||c|c|c|c|c|} \hline
    Flow Type    
         & Benign
            & HTTPFlood
                & SlowrateDoS
                    & UDPFlood
                        & Anomalous  \\ \hline 
    \# of images
         & 5081
            & 1497
                & 777
                    & 4865
                        & 706 \\ \hline
    \end{tabular}
    }
    \caption{Summary of 5G network traffic images.}
    \label{tab:5g-dataset}
\end{table}

\begin{figure*}[]

\begin{subfigure}{.245\linewidth}
  \includegraphics[width=\linewidth]{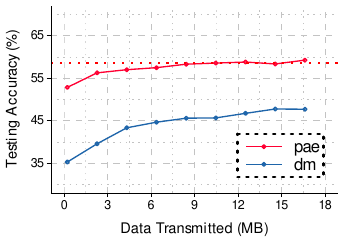}
  \caption{IPC = 10, \# of clients = 5}
  \label{fig:05_10}
\end{subfigure}\hfill % <-- "\hfill"
\begin{subfigure}{.245\linewidth}
  \includegraphics[width=\linewidth]{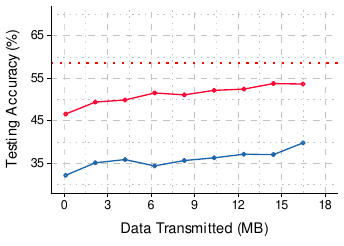}
  \caption{IPC = 10, \# of clients = 20}
  \label{fig:20_10}
\end{subfigure}
\begin{subfigure}{.245\linewidth}
  \includegraphics[width=\linewidth]{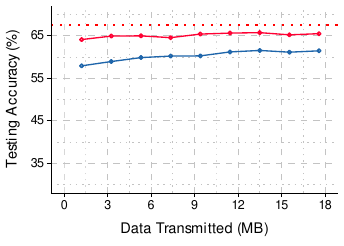}
  \caption{IPC = 50, \# of clients = 5}
  \label{fig:05_50}
\end{subfigure}\hfill % <-- "\hfill"
\begin{subfigure}{.245\linewidth}
  \includegraphics[width=\linewidth]{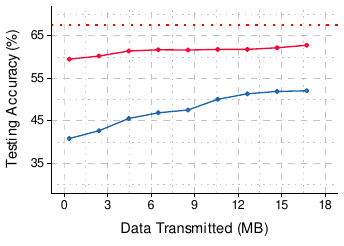}
  \caption{IPC = 50, \# of clients = 20}
  \label{fig:20_50}
\end{subfigure}
\caption{
Testing accuracy vs. data transmitted per client across different parameter settings. 
The dashed red line corresponds to the classification accuracy of partition-and-expand distribution matching in the central model.}
\label{fig:poc}
\end{figure*}

\paragraph*{Datasets}

For benchmark testing, we conduct experiments on four standard image classification datasets: MNIST \cite{lecun1998gradient}, FMNIST \cite{xiao2017fashion}, CIFAR10 \cite{krizhevsky2009learning} and SVHN \cite{netzer2011reading}. 
MNIST consists of 60,000 binary images of handwritten digits.
There are 10 classes in total.
FMNIST is a dataset of 70,000 binary images of fashion products with 10 classes.
CIFAR10 is a selection of 50,000 small color images separated into  10 classes.
SVHN is a dataset of 600,000 color images of street sign numbers with 10 classes.

For attack detection on 5G mobile networks, we adopt the 5G-NIDD dataset \cite{samarakoon20225g}, a comprehensive benchmark dataset for 5G attack detection.
The dataset is labeled and constructed through the creation of benign and malicious traffic profiles on a functional 5G test network. 
The data was collected in an environment comprised of 2 base stations connected to an attacker node and a set of benign traffic-generating devices.
Malicious traffic is generated through either DoS or port scan attacks.
The network captures are processed into CSV files containing the network flows and their associated features.
Each flow is classified as either benign or as belonging to one of 8 different attacks.
The classes of attack include:
\begin{itemize}
    \item DoS: HTTP Flood, ICMP Flood, SYN Flood, Slowrate DoS and UDP Flood.
    \item Port Scan: UDP Scan, Syn Scan, and TCP connect Scan.
\end{itemize}
Each flow is marked with 44 features.
Rows from the CSV file are combined into batches and transformed into 64x64 black/white images.
In total, 12,295 images were created. 
As the classes ICMP Flood, SYN Flood, SYN Scan, TCP Connect Scan, and UDP Scan contained an insufficient number of images, they were combined into an umbrella class of `anomalous' traffic.
The resulting dataset is summarized in Table \ref{tab:5g-dataset}

\begin{table*}[]
    \centering
    \resizebox{\columnwidth}{!}{%
    \begin{tabular}{|c||ccc|ccc|ccc|ccc|} \hline \\[-1.0em]
    Dataset
        & 
            & MNIST
                &
        & 
            & FMNIST
                &
        & 
            & CIFAR10
                &  
        & 
            & SVHN
                &         \\ \hline  \\[-1.0em]
    Skew
        & $\beta = 0.1$
            & $\beta = 0.3$
                & $\beta = 0.5$
        & $\beta = 0.1$
            & $\beta = 0.3$
                & $\beta = 0.5$
        & $\beta = 0.1$
            & $\beta = 0.3$
                & $\beta = 0.5$ 
        & $\beta = 0.1$
            & $\beta = 0.3$
                & $\beta = 0.5$       \\ \hline \hline \\[-1.0em]
    \textsf{FedD3} \cite{song2023federated}
        &  86.70 
            & 88.91
                & 88.36
        &  64.35 
            & 74.13
                & 75.15
        & 38.54  
            & 39.37
                &  40.53
        & 65.11
            & 66.00
                &  65.62       \\ \hline \\[-1.0em]   
    \textsf{DOSFL} \cite{zhou2020distilled}
        & 77.97   
            & 83.24
                & 91.73
        & 64.25  
            & 72.01
                & 80.83
        & 43.92  
            & 47.08
                & 56.62  
        & 67.94
            & 69.05
                &  69.91       \\ \hline \\[-1.0em]            
    \textsf{DENSE} \cite{zhang2022dense}
        & 66.61  
            & 76.48
                & 95.82
        & 50.29  
            & 83.96
                & 85.94
        & 50.26  
            & 59.76
                &  {62.19}
        & 55.34
            & {79.59}
                & {80.03}        \\ \hline \\[-1.0em]
    \textsf{LocalDM}
        & 96.10 
            & 96.93
                & 97.17
        & 84.18
            & 84.5
                & 84.24
        & 52.93
            & 52.17
                &   54.22
        & 70.04
            & 70.69
                &   71.49      \\ \hline \hline \\[-1.0em]
    \textsf{CollabDM}
        & {97.72}
            & \textbf{97.82}
                & {97.83}
        & {85.43}
            & \textbf{86.51}
                & {86.71}
        & {57.97} 
            & 59.36
                &   {60.21}
        & {74.35}
            & 74.66
                & 75.57       \\ \hline \\[-1.0em]
    \textsf{CollabDM-pae}
        & \textbf{97.78}
            & 97.80
                & \textbf{98.07}
        & \textbf{86.19}
            & 86.31
                & \textbf{86.91}
        & \textbf{63.91}
            & \textbf{64.67}
                &  \textbf{64.50}
        & \textbf{85.83}
            & \textbf{86.44}
                & \textbf{86.53}       \\ \hline            
    \end{tabular}
    }
    \caption{Accuracy of different methods across $\beta =\{0.1,0.3,0.5\} $ on different datasets.
    IPC = 50 for the distillation methods. }
    \label{tab:het_learning}
\end{table*}

\paragraph*{Data Partition}
To simulate real-world applications and distributed learning environments, we use the Dirichlet distribution to generate a non-IID data partition among clients \cite{yurochkin2019bayesian,zhang2022dense}. 
In particular, we sample $p_k~\sim \textsf{Dir}(\beta)$ and allocate a $p_k^i$ proportion of the data of class $k$ to the client $i$.
We can change the degree of imbalance by varying the parameter $\beta$. A small $\beta$ generates a highly skewed partition.

\paragraph*{Model Architecture}

Unless otherwise specified, all synthetic data are distilled using embeddings from the convolutional neural network (ConvNet) architecture used by Zhao \textit{et al.} \cite{zhao2023dataset}. 
For classification accuracy, the learned synthetic sets are used to train randomly initialized ConvNets, which are subsequently used to perform classification tasks on real data.
The default ConvNet includes three repeated convolutional blocks, and each block involves a 128-kernel convolution layer, instance normalization layer \cite{ulyanov2016instance}, ReLu activation function \cite{nair2010rectified}, and average pooling.
In each experiment, we learn one synthetic set and use it to test 20 randomly initialized networks. We repeat each experiment 5 times and report the mean testing accuracy of the 100 trained networks. 
In addition, to test the transferability of the synthetic data, do cross-architecture experiments in Section~\ref{subsec:5g}. 
Following Zhao \textit{et al.} \cite{zhao2023dataset},
we evaluate our method on four different architectures, including ConvNet, AlexNet \cite{krizhevsky2012imagenet}, VGG11 \cite{simonyan2014very} and ResNet18 \cite{he2016deep}. 
In this setting, we learn the synthetic set on one network architecture and use the resulting set to train networks with different architectures.
The ability to train different network architectures on the same synthetic set is an advantage collaborative data distillation has over traditional federated learning.

\paragraph*{Comparison Methods}

We evaluate two versions of \collabdm.
These include the standard version, outlined in Algorithm \ref{alg:CollabDM}, and the optimized version, outlined in Section \ref{sec:opt}, that utilizes the partition and expand technique \cite{zhao2023improved}.
For the optimized version, denoted \textsf{CollabDM-pae}, clients also employ partition-and-expand for the local distillation step.

For benchmarking, we compare \collabdm against four baselines.
The first, named \textsf{LocalDM}, is based on naive virtual learning, where clients distill their data \textit{independently} of each other, and the server uses this data for model training without refining it over a global objective.
By using distribution matching as the local objective,
we can assess the gain in classification accuracy achieved through the additional steps in \collabdm to refine global synthetic data.
We also against compare two prior works (FedD3 \cite{song2023federated} and DOSFL \cite{zhou2020distilled})  that use naive virtual learning with the meta-learning distillation objective.  
DOSFL is implemented with an optimization from Feng \textit{et al.} \cite{feng2023embarrassingly} to update the method.

In addition, we evaluate \textsf{Collab} against \textsf{DENSE} \cite{zhang2022dense}, a state-of-the-art technique for one-shot federated learning.
\textsf{DENSE} creates a global data generator and uses the data generator to train models on the server.
For a fair comparison, we only include data synthesizing techniques.
Thus, one-shot averaging methods, such as \cite{su2023one}, are not included.
These methods can provide a good classification model for training a \textit{single} network. 
However, unlike data synthesizing techniques, they cannot be used for additional applications such as data sharing, neural architecture search, and continual learning.

\paragraph*{Training parameters}
Following Zhao \textit{et al.} \cite{zhao2023dataset}, the learning rate for local distillation is set to 1.0.
The number of training iterations is set to 1000, compared to 20,000 in prior work.
This is to reduce resource consumption on the client side, with each client taking 2-3 minutes to complete their portion of the algorithm.
We also use a larger batch size of 512 for embeddings on real data. 
This will support faster convergence at the server and reduce the amount of data transmitted over the network.
In addition, the learning rate for synthetic data at the server is set to 10.
Again, this is designed to encourage faster convergence.
During the experiments, we measure classification accuracy at every 50 iterations.
This allows the trade-off between data transmission and classification accuracy to be observed.

\subsection{Benchmark Image Data}

\subsubsection*{Parameters}

We begin by looking at the impact of key parameters on classification performance.
The parameters under consideration are images-per-class, the number of clients, and the number of iterations performed for global distillation.
This experiment will also provide a proof-of-concept.
That is, it will demonstrate that the global distillation steps improve classification accuracy within a distributed setting. 
For this section, the dataset is CIFAR10, and data is distributed IID across clients.

The results of the experiment are displayed in Figure~\ref{fig:poc}.
The number of iterations is expressed as the amount of data transmitted across the channel.
The results demonstrate that increasing the amount of information transmitted increases the classification accuracy of the synthetic set.
However, diminishing returns are experienced, and the largest increases in accuracy occur during the early iterations. 
As expected, testing accuracy is inversely proportional to the number of clients.
For example, as observed in Figures \ref{fig:05_10} and \ref{fig:20_10}, the classification accuracy for \textsf{CollabDM-pae} drops from $59\%$ to $53\%$ as the number of clients increases from 5 to 20.
Increasing the number of images-per-class only has a small impact on the amount of data transmitted, while significantly increase testing accuracy.
For example, for \textsf{CollabDM-pae}, as observed in Figures \ref{fig:05_10} and \ref{fig:05_50}, at 10 images-per-class 4.3MB of data achieves $57\%$ accuracy and at 50 images-per-class 3.27MB of data achieves $65\%$ accuracy. 
Notably, across all four settings, the partition-and-expand technique provides a significant increase in classification accuracy.   

\begin{figure}[t]
    \centering
    \includegraphics[width=0.3\columnwidth]{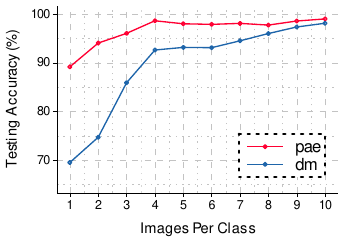}
    \caption{The impact of images-per-class on testing accuracy for 5G network traffic data.}
    \label{fig:5g-ipc}
\end{figure}

\subsubsection*{Heterogeneous One-Shot Learning}

The evaluation of \collabdm in heterogeneous one-shot learning is listed in Table~\ref{tab:het_learning}.
The number of iterations at the server is set to 200. 
This is equivalent to at most 9MB of data being sent per client.
The level of heterogeneity is controlled by the parameter $\beta$.
Most notably, all three distribution matching techniques demonstrate remarkable robustness to data heterogeneity, with all three outperforming the state-of-the-art method for $\beta=0.1$.
The surprising performance of \textsf{LocalDM} provides evidence that distribution matching techniques are well suited to collaborative data distillation and distributed learning settings.

\textsf{CollabDM} significantly outpeforms both \textsf{FedD3} and \textsf{DOSFL}, the two techniques that perform local meta-learning distillation. 
Song \textit{et al.} originally test \textsf{FedD3} in a setting with a small dataset and a large number of clients (50000 data points spread across 500 clients) \cite{song2023federated}.
In their experiments, they rely on 500 clients distilling 10 images each, which leads to a synthetic set at the client of 5000 images.
This is a brute approach that is effective at reducing communication when the scope is limited to just federated learning.
However, under our objective, Problem \ref{prob:cdd}, the method  performs poorly when asked to distill a \textit{compact} synthetic set of 50 images per class.
For completeness, as \collabdm sends additional data, we also compared \textsf{FedD3} and \collabdm when the volume of communication is equal. 
With each client limited to 9MB (which equates to $600$ images per client for \textsf{FedD3}), \collabdm outperforms \textsf{FedD3} on all tests, including an improvement of 5 percentage points on CIFAR10.

DOSFL uses \textit{soft resets}, where each training model is sampled from the parameters of the server's model, to overcome data heterogeneity \cite{zhou2020distilled}.
While soft resets outperform traditional random initializations on non-IID data, they are not robust against varying degrees of skew.
This demonstrates the need for a global distillation objective to regularize learning.

\textsf{CollabDM-pae} outperforms the state-of-the-art \textsf{DENSE} in \textit{all} experiments, with notable improvements for highly skewed data partitions ($\beta=0.1$).
For example, on the SVHN dataset, for $\beta=0.1$, \textsf{CollabDM-pae} improves over \textsf{DENSE} by 30 percentage points.

\subsection{5G Attack Detection}
\label{subsec:5g}

Attack detection on 5G network data is a motivating application for our technique.
5G networks are decentralized by design and cater to a range of verticals.
It is a setting in which data generation is innately distributed and heterogeneous.
In our test setting, data collection is split across two base stations, which act as the clients in \collabdm.
For all experiments, the amount of data transmitted is limited to $9$MB per client.

We first look at the impact of the number of images per class on attack classification.
The results are presented in Figure~\ref{fig:5g-ipc}.
Remarkably, at just 1 image-per-class,  \textsf{CollabDM-pae} achieves $89\%$ testing accuracy.
This represents the distillation of 12,995 images of network traffic into 5 informative images, which, with the partition-and-expand technique, contain enough information to allow the network to not only distinguish between benign and malicious flows but also classify concrete attacks.
This suggests that different attacks have highly distinct profiles.
In addition, at just 10 images-per-class, \textsf{CollabDM-pae} achieves peak testing accuracy at $99\%$.

To verify the generalizability of the global synthetic sets, we conduct cross-architecture experiments.
The results are presented in Table~\ref{tab:5g-cross-arch}. 
Synthetic data is learned on one architecture and evaluated on a separate architecture.
Each synthetic set contains 10 images per class.
Results indicate that the synthetic sets generalize well, with, at worst, only a small drop in accuracy when moving to new architectures.
These results promote the use of data distillation for data sharing in 5G networks, with very small global synthetic sets available for machine learning applications at different locations in the network.

\begin{table}[t]
    \centering
    \resizebox{0.5\columnwidth}{!}{
  \begin{tabular}{|c||c|c|c|c|} \hline
    \backslashbox{Train}{Test} 
        & ConvNet 
            & AlexNet
                & VG11
                    & ResNet \\\hline \hline
       ConvNet 
        & 98.84 
            & 96.49 
                & 98.18
                    & 97.60 \\ 
        AlexNet 
        & 96.55 
            & 94.21 
                & 95.87
                    & 94.19 \\ 
        VG11 
        & 89.95 
            & 86.31 
                & 89.33
                    & 91.30 \\ 
        ResNet 
        & 95.71 
            & 93.22 
                & 95.05
                    & 94.16 \\ \hline
    \end{tabular}}
    \caption{Cross-architecture evaluation for 5G network traffic data.}
    \label{tab:5g-cross-arch}
\end{table}

%% file: sections/06_conclusion.tex
We have presented a novel algorithm for data distillation in distributed settings.
The algorithm supports the distillation of a synthetic set that matches the global data distribution and requires only a single round of communication between clients and the central server.
Experiments demonstrate that learned synthetic sets are robust to heterogeneous data partitions and comfortably outperform the state-of-the-art approach. 
In addition, our work is motivated by a new application for data distillation: attack detection in 5G mobile networks.
Experiments exhibit that distillation techniques effectively capture the information in both benign and malicious traffic profiles.